\documentclass[final,twocolumn]{elsarticle}

\usepackage{lineno,hyperref}
\usepackage{color}
\usepackage{algorithm}
\usepackage{amsmath}
\usepackage{amsfonts}
\usepackage{nomencl}
\makenomenclature
\modulolinenumbers[5]

\journal{Mechatronics}

\begin{document}
	
	\begin{frontmatter}
		
		\title{Coordinated Motion Control and Event-based Obstacle-crossing for Four Wheel-leg Independent Motor-driven Robotic System\tnoteref{mytitlenote}}
		
		\tnotetext[mytitlenote]{This work was supported in part by the National Key Research and Development Project of China under Grant 2019YFC1511401 and the National Natural Science Foundation of China under Grant 51675041 and 61773060.}
		
		\author[mymainaddress,myseaddress]{Dongchen Liu}
		
		\author[mymainaddress,myseaddress]{Junzheng Wang\corref{mycorrespondingauthor}}
		\cortext[mycorrespondingauthor]{Corresponding author}
		\ead{Wangjz@bit.edu.cn}
		
		\author[mymainaddress,myseaddress]{Shoukun Wang}
		
		\address[mymainaddress]{The National Key Laboratory of Complex System Intelligent Control and Decision, School of Automation, Beijing Institute of Technology, Beijing, 10081, China}
		\address[myseaddress]{The Ministry of Industry and Information Technology Key Laboratory of Servo Motion System Drive and Control, School of Automation, Beijing Institute of Technology, Beijing, 10081, China}
		
		\begin{abstract}
			
			This work investigates the coordinated motion control and obstacle-crossing problem for a four wheel-leg independent motor-driven robotic system via a model predictive control (MPC) approach based on an event-triggering mechanism. The modeling of a wheel-leg robotic control system with a dynamic supporting polygon is organized. The system dynamic model is 3 degrees of freedom ignoring the pitch, roll, and vertical motions. The single wheel dynamic is analyzed considering the characteristics of the motor-driven and the Burckhardt nonlinear tire model. As a result, an over-actuated predictive model is proposed with the motor torques as inputs and the system states as outputs. As the supporting polygon is only adjusted at certain conditions, an event-based triggering mechanism is designed to save hardware resources and energy. The MPC controller is evaluated on a virtual prototype as well as a physical prototype. The simulation results guide the parameter tuning for the controller implementation in the physical prototype. The experimental results on these two prototypes verify the efficiency of the proposed approach.
			
		\end{abstract}
		
		
	\end{frontmatter}
	
	\section{Introduction}
	
	Four-wheel independent motor-drive (4WIMD) vehicles play an important role in wheeled intelligent systems, due to their advantages in driveability and {\color{blue}energy-saving} \cite{JIN2015286,4WIMDb,XU201995,ISATransLIU2020}. Coordinated motion control of these systems, which is an essential point for satisfying handling and stability performance, has been a popular topic in recent decades\cite{LEE2018157,3,4,6}. Different achievements about the coordinated motion control strategy have been achieved, e.g., proportional-integral-derivative control \cite{PI}, sliding mode control \cite{DU2021102484}, adaptive control \cite{HUANG201860}, robust control \cite{adaptive}, and model predictive control (MPC) \cite{MPC}. {\color{blue}In particular, the MPC control scheme, which is a closed-loop strategy that adds feedback to typical open-loop optimal control solutions, suits nonlinear systems with state quantities constraints \cite{Guo2020A}}. In \cite{MPC2}, a novel visual servo-based model predictive control method is developed to steer a wheeled mobile robot moving in a polar coordinate toward the desired target. A tracking control system with {\color{blue}an explicit nonlinear} model predictive control scheme is presented in \cite{MPC3} for {\color{blue}electric-driven vehicles with in-wheel motors. A key point of the MPC controller design is the predictive system modeling.}
	
	
	There exist various effective modeling approaches to meeting the Ackermann steering mechanism. In \cite{Acherman1}, a synchronization control approach is proposed with a two-wheeled-driven electric vehicle regarding two wheels as followers. The synchronization action is achieved by using a general fictitious master, and the Ackermann principle is used to compute an adaptive desired wheel speed, resulting in a robust technique. As the existence of the nonlinear coupling effects in the Ackermann steering mechanism, neural networks have been applied to estimate the vehicle speed \cite{Acherman2}. In \cite{4WIMEs}, the design and evaluation of electrical differential (ED) are discussed for over actuated electric vehicles independently driven by in-wheel motors. From the comparison of three patterns of ED, almost the same vehicle performances in terms of the sideslip angle, yaw rate, and trajectory tracking have been achieved under normal driving maneuvers. The stabilization of control architecture based on the Ackermann steering mechanism is limited under adverse driving conditions when the wheels are slipping or adjusting the relative position. Moreover, another kind of widely used method targets the adhesion or slip ratio control of each wheel. {\color{blue}In \cite{wslip1}, based on the measurements from wheel encoders, an inertial measurement unit, and a global positioning system, the joint wheel-slip, as well as the estimation about vehicle motion, are considered.} As automated vehicles and advanced driver-assistance systems {\color{blue}benefit from} vehicle trajectory prediction to understand the traffic environment and perform tasks, an integrated vehicle trajectory prediction method, which is based on combining physics and maneuver approaches, is proposed in\cite{wslip2}. Actually, in terms of the path-following and safety problem, a nonlinear coordinated motion controller is proposed in \cite{wslip3} with the framework of the triple-step method that potentially benefits the engineering-oriented implementation for intelligent vehicles. However, the real-time observation of small slip rates and the identification of optimal slip rates under complex driving conditions are difficult to achieve. With the shortcomings of the methodologies investigated above, the improvement of modeling for the 4WIMD systems is significant for the MPC controller design.
	
	In most of the studies cited above, the supporting polygon is changeless, and the vehicles only can bypass obstacles from the side. Recently, as diverse wheeled robots have realized the flexible adjustment of the wheels, the systems can cross obstacles by adjusting the supporting polygon to shorten the journey and strengthen the terrain adaptability \cite{9003193,8967935,xu2020adaptive}. {\color{blue}The relative position changes between wheels result in a nonlinear time-varying model for the dynamics of the robotic system. The rapidly increasing number of calculations place extra demands on hardware resources and battery life which are limited on robotic systems. The MPC-based coordinated motion control data for this kind of system are still scarce, and development efforts are still necessary.}
	
	In this work, we propose a thorough study of a coordinated motion control problem for a {\color{blue}four-wheel-leg} robotic system. As the wheels can steer and adjust position independently, the system has different dynamic characteristics from 4WIMD vehicles. Motivated by the Ackermann steering mechanism, the wheel steering angles are calculated according to the reference trajectory obtained from the environment awareness system. {\color{blue}The steering angle for each wheel is along its own steering axis and the desired paths for all wheels are around the same point, which meets the characteristics of our independent driven robotic system.} Furthermore, a predictive model for this kind of system is proposed with the motor torques as inputs and the system states as outputs. In practice, the adjustments only occur in certain conditions, exemplified by deep pits or isolated obstacles. Therefore, the event-based triggering mechanism is suitable. Pioneered by the works in \cite{EVENT1,EVENT2,EVENT3}, numerous results have been obtained in event-based control and estimation \cite{EVENT4,HUANG2021131}. Li and Shi \cite{EVENT5} combine the event-triggered theory with the MPC strategy for continuous-time nonlinear systems with disturbances. In \cite{EVENT6}, the event-triggered extended state observer (ET-ESO) is designed for a continuous-time nonlinear system with system uncertainty and outside disturbance. As for motion control of robotic systems, a behavior-based switch-time MPC controller for mobile robots in \cite{EVENT7} meets the characteristics of the coordinated motion control of wheeled robots with a dynamic supporting polygon. Admittedly, relevant literature has rarely been seen about the event-based modeling for coordinated motion control of wheeled robotic systems. We consider the coordinated motion control as different behaviors including steering and adjusting supporting polygon with the switch time determined by an event-based triggering mechanism. The main contributions of our work are as follows:
	\begin{itemize}
		\item A developed model considering both the Ackermann steering mechanism and the wheel adhesion for the four wheel-legs independent motor-driven systems is proposed. The model is 3 degrees of freedom and {\color{blue}is established} by the characteristics of wheel motion with a dynamic supporting polygon.
		\item An event-based triggering mechanism is designed based on the information from the environment awareness system. The improved MPC controller is discussed in theory and evaluated on a virtual prototype and a physical prototype. The experimental results verify the effectiveness of the proposed approach.
	\end{itemize}
	
	The rest of this paper is organized as follows. In section II, the model of four wheel-leg independent motor-driven robotic systems is proposed. In section III, an event-based MPC coordinated motion controller is shown in detail. Simulation and experiment results are presented in section IV, and the paper concludes with a brief discussion in section V.
	
	\section{System Model and Problem Formulation}
	
	For the controller design of a four wheel-leg independent motor-driven robot, a dynamic model is required. {\color{blue} With the reference trajectory of the system as input, the wheel steering angles can be calculated by Ackermann steering geometry \cite{Acherman1,Acherman2,4WIMEs}, such that the desired paths for all wheels are around the same point. The steering angle for each wheel is along its own steering axis, which meets the characteristics of our independent driven robotic system. However, the dynamics of the robotic system and each wheel are required for this kinematic model under adverse driving conditions when the wheels are slipping.} 
	
	\subsection{System Dynamic Model}
	
	\begin{figure}[t]
		\centering
		\includegraphics[width=0.85\hsize]{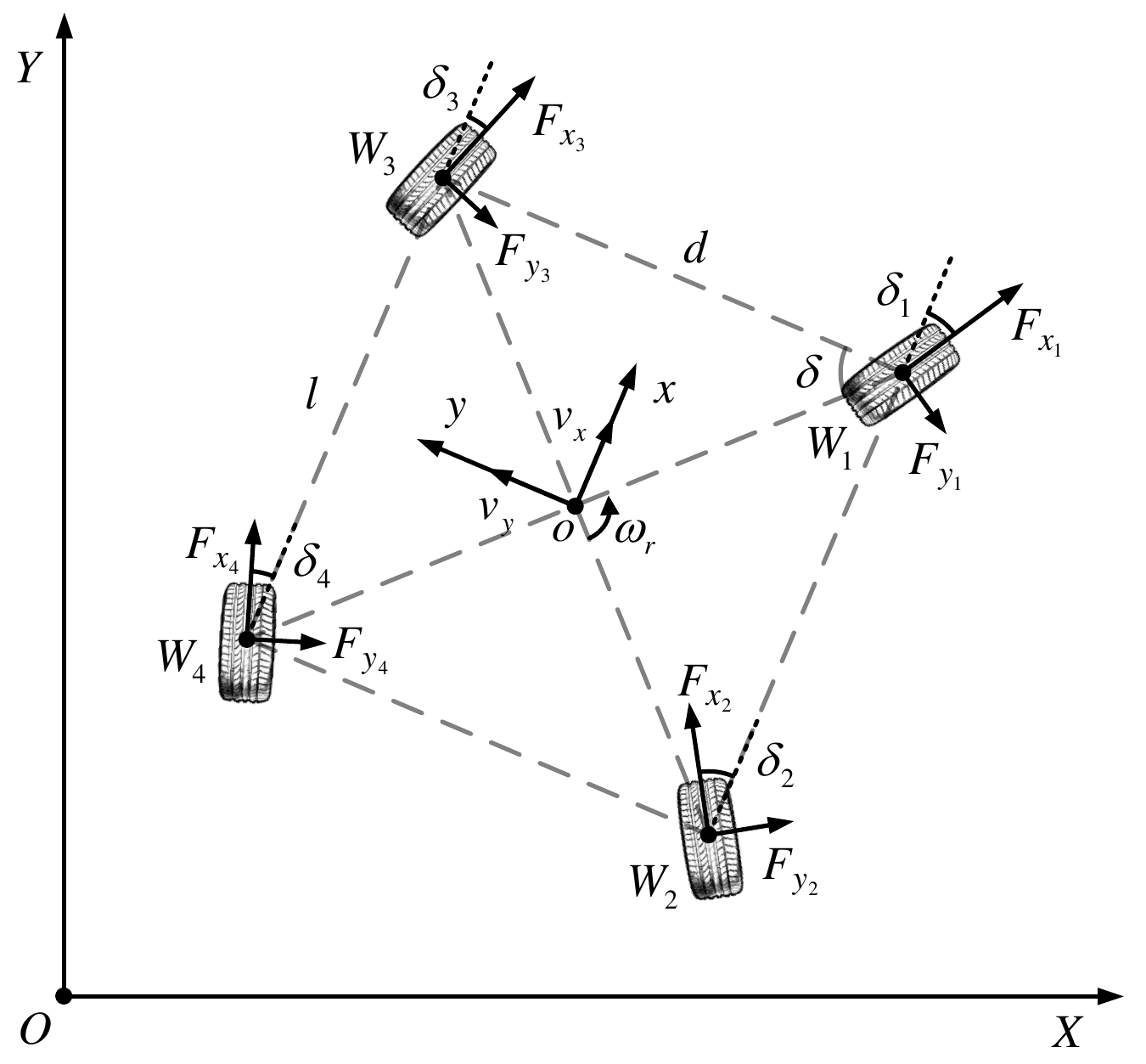}
		\caption{Schematic diagram of the robotic system.}
		\label{fig1}
	\end{figure}
	
	The schematic diagram of a four wheel-leg independent motor-driven robotic system in  the longitudinal, lateral, and yaw directions is shown in Fig.~\ref{fig1}. The wheels are numbered as $W_i$, where $i=1,2,3,4$. To express the ratio between the wheelbase $l$ and track width $d$, $\delta$ is defined as
	\begin{equation}
	\label{eqsm0}
	\begin{aligned}
	\delta=\arctan\left( \frac{l}{d} \right).
	\end{aligned}
	\end{equation}
	{\color{blue} The tire forces in longitudinal, lateral and vertical directions are presented by $[F_{x_i}, F_{y_i}, F_{z_i}]$}. The origin $o$ of the dynamic coordinate system $\Sigma_b$ is fixed on the system coinciding with the center of gravity (COG), the $x-$axis is the longitudinal axis of the system (the forward direction is positive), the $y-$axis is the lateral axis of the system (the right-to-left direction is positive). The steering angle between the longitudinal direction of wheel $W_i$ and $x-$axis is $\delta_i$. In the geodetic coordinate system $\Sigma_n$, $X-$axis positive direction is due east, and $Y-$axis positive direction is due north. This model is 3 degrees of freedom (DOF) ignoring the pitch, roll, and vertical motions of the system. It is assumed that the mechanical properties of the wheels are the same.
	
	The motion of the system includes translation $v_t=[v_x,v_y]$ and rotation $\omega_r$. In this paper, the velocity vectors are uniformly mapped to a velocity vector $v=[v_t, \omega_r]^T$. Considering the Ackermann steering mechanism, we have
	\begin{equation}
	\label{eqsm1}
	\begin{aligned}
	&\delta_1=-\delta_2=\arctan\left(\frac{lK}{2-dKC_d}\right),\\
	&\delta_3=-\delta_4=\arctan\left(\frac{lK}{2+dKC_d}\right),
	\end{aligned}
	\end{equation}
	where $K$ is the turning curvature which is inversely proportional to the turning radius, and $C_d$ is the turning direction. As the reference trajectory is described by ${x}^\ast=[P_x^\ast(t), P_y^\ast(t), P_\theta^\ast(t)]^T$, it holds that
	\begin{equation}
	\label{eqsm2}
	\left\{
	\begin{aligned}
	&K=\frac{|\dot{P}_x^\ast(t)\ddot{P}_y^\ast(t)-\ddot{P}_x^\ast(t)\dot{P}_y^\ast(t)|}{\left[ \dot{P}_x^\ast(t)^2+\dot{P}_y^\ast(t)^2 \right]^{\frac{3}{2}}},\\
	&C_d=\left\{
	\begin{array}{cc}
	\!\!1,  & P_\theta^\ast(t)>0, \\
	\!\!0,  & P_\theta^\ast(t)=0, \\
	\!\!-1, & P_\theta^\ast(t)<0. \\
	\end{array}
	\right.
	\end{aligned}
	\right.
	\end{equation}
	From dynamic analysis, considering the mass $m$ and inertia $I$ of the wheeled system, we have
	\begin{equation}
	\label{eqsm3}
	\!\!\!\left\{\!\!
	\begin{aligned}
	m\dot{v}_x      \!\!=& \sum_{i=1}^4 \left( F_{x_i}\cos\delta_i + F_{y_i}\sin\delta_i \right) + m\omega_r v_y,\\
	m\dot{v}_y      \!\!=& \sum_{i=1}^4 \left( F_{x_i}\sin\delta_i + F_{y_i}\cos\delta_i \right) - m\omega_r v_x,\\
	I\dot{\omega}_r \!\!=& \frac{\sqrt{l^2\!\!+\!d^2}}{2} \!\!\sum_{i=1}^4 \!\!\left( F_{\!x_i}\!\cos\left( \delta_i\!\!-\!\delta \right) \!+\! F_{\!y_i}\!\sin\left( \delta_i\!\!-\!\delta \right) \!\right).\\
	\end{aligned}
	\right.
	\end{equation}
	The wheel sideslip angles $\alpha_i$ are calculated by
	\begin{equation}
	\label{eqsm4}
	\left\{
	\begin{aligned}
	\alpha_1 &= \delta_1 - \frac{2v_y + l\omega}{2v_x - d\omega},\\
	\alpha_2 &= \delta_2 - \frac{2v_y - l\omega}{2v_x - d\omega},\\
	\alpha_3 &= \delta_3 - \frac{2v_y + l\omega}{2v_x + d\omega},\\
	\alpha_4 &= \delta_4 - \frac{2v_y - l\omega}{2v_x + d\omega}.\\
	\end{aligned}
	\right.
	\end{equation}
	
	\subsection{Single Wheel Dynamic Analysis}
	
	\begin{figure}[t]
		\centering
		\includegraphics[width=1.0\hsize]{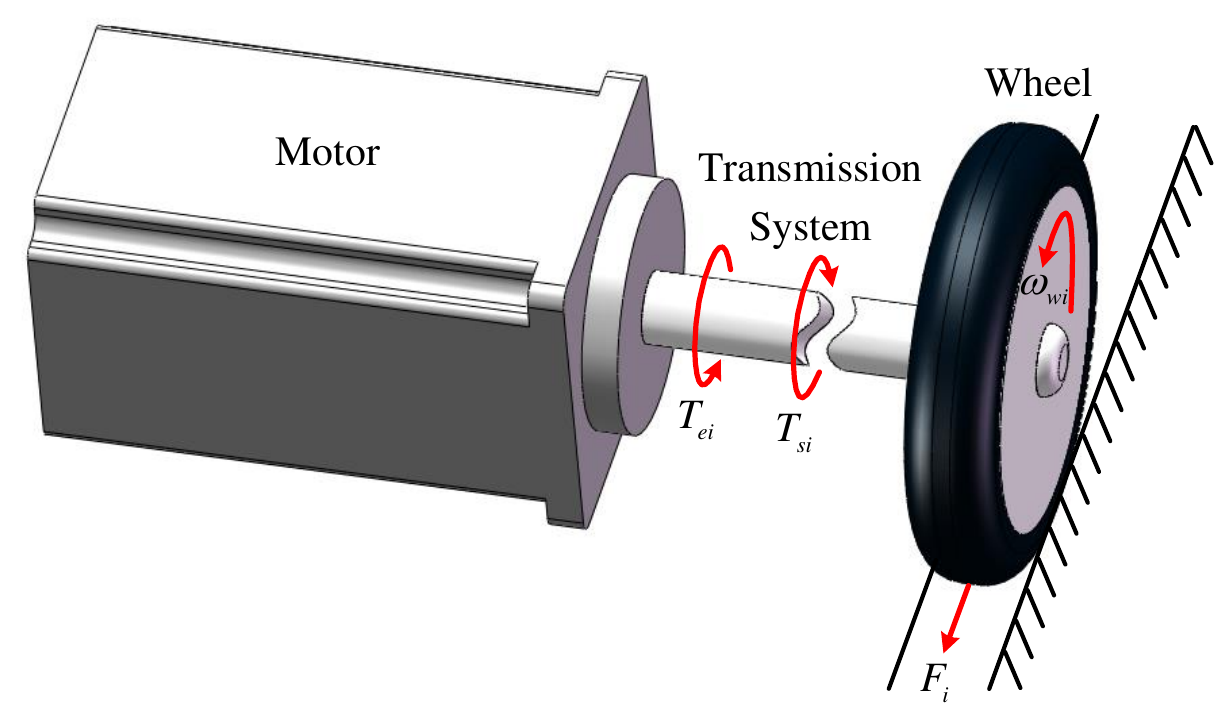}
		\caption{Single wheel dynamics}
		\label{fig2}
	\end{figure}
	
	For a four wheel-leg independent motor-driven robotic system, each wheel is driven by a motor. A transmission system is adopted to transfer the electromagnetic driving torque $T_{e_i}(u_i)$ from the motor to the wheels. In practice, the transmission requires multiple components such as the gearbox, torque shaft, and suspension mechanism, such that the resistance $T_{s_i}$ exists. The dynamic of a single wheel is simplified as shown in Fig.~\ref{fig2}, and characterized by
	\begin{equation}
	\label{eqw1}
	\begin{aligned}
	T_{e_i}(u_i) - T_{s_i} - F_{x_i}r= J_\textit{w} \dot{\omega}_{\textit{w}_i},
	\end{aligned}
	\end{equation}
	where $\omega_{\textit{w}_i}$, $J_\textit{w}$ and $r$ are the rotation velocity, inertia, and radius of the wheel respectively. Referring to \cite{Mfs}, $T_{e_i}(u_i)$ is approximated by $k_iu_i$. However, the tire is highly nonlinear and the Burckhardt tire model is suitable for controller design to describe its nonlinearity \cite{MPC}. The composition of the tire longitudinal and lateral forces is calculated by
	\begin{equation}
	\label{eqw2}
	\left\{
	\begin{aligned}
	&F_{t_i}=F_{z_i}c_1(1-e^{c_2s_{iRes}})-c_3s_{iRes},\\
	&F_{x_i}=\frac{\lambda_i}{s_{iRes}}F_{t_i},\\
	&F_{y_i}=\frac{\tan\alpha_i}{s_{iRes}}F_{t_i},\\
	\end{aligned}
	\right.
	\end{equation}
	where $F_{t_i}$ is the composition of the longitudinal and lateral tire forces, $F_{z_i}$ is the vertical load of the wheels and $[c_1, c_2, c_3]$ are parameters for the dry asphalt road. The combined friction efficient satisfies
	\begin{equation}
	\label{eqw3}
	\begin{aligned}
	s_{iRes}=\sqrt{\lambda_i^2+{\tan\alpha_i}^2}
	\end{aligned}
	\end{equation}
	with the wheel longitudinal slip is calculated by
	\begin{equation}
	\label{eqw4}
	\begin{aligned}
	\lambda_i=\frac{\omega_{\textit{w}_i}r-v_x}{\max(\omega_{\textit{w}_i}r,v_x)}.
	\end{aligned}
	\end{equation}
	
	\subsection{Problem Formulation}
	
	The supporting polygon changes when the robot crosses an isolated obstacle. From this point, an event-based triggering mechanism is effective with the triggering condition $\Gamma(t)$ proposed as
	\begin{equation}
	\label{eqpf1}
	\Gamma(t)=\left\{
	\begin{aligned}
	&1,  ~~\textrm{if} ~d_s\in(d_0, d_0+\Delta d_{\textrm{max}}),\\
	&0,  ~~\textrm{otherwise},\\
	\end{aligned}
	\right.
	\end{equation}
	where $d_s$ is the obstacle width which can be obtained by the environment awareness system, $d_0$ is the initial lateral distance of wheels, and $\Delta d_{\textrm{max}}$ is the maximum stretch width limited by the workspace of the legs. Taking $x=v_t^T$ and $u=[u_{1}, u_{2}, u_{3}, u_{4}]^T$ as the states and control inputs of the system model, the control system satisfies
	\begin{equation}
	\label{eqpf2}
	\begin{aligned}
	\dot{x}=g(x,u,\Gamma),
	\end{aligned}
	\end{equation}
	where $g(x,u,\Gamma)$ is constructed from (\ref{eqsm3}) to (\ref{eqw4}). This model includes the coupling of the robotic system dynamics and single wheel nonlinearity.	In this work, we mainly focus on the coordinated motion control for the four-wheel independent motor-driven robotic system with the following problems:
	\begin{itemize}
		\item Considering the system control model in (\ref{eqpf2}), can we propose an coordinated motion controller based on MPC for a four wheel-leg independent motor-driven robotic system with a dynamic supporting polygon?
		\item As the complexities of the dynamic character, can we propose a event-based framework to simplify the control problems? With the event-triggering condition $\Gamma(t)$ come into concern, how do we achieve supporting polygon switch with stable motion control?
	\end{itemize}
	
	\section{Event-based MPC Controller of Coordinated Motion Control}
	
	In this section, the problems stated above are investigated. Consider the event-based supporting polygon adjustment of the wheeled system and define the behaviors as $\alpha(x(t),\Gamma(t))$. Then, the control system in (\ref{eqpf2}) is expressed as
	\begin{equation}
	\label{eqmt1}
	\begin{aligned}
	\dot{x}=f(x,u,\alpha(x,\Gamma)).
	\end{aligned}
	\end{equation}
	As the number of the control inputs is more than the number of the model states, the control problem is over-actuated. More specifically, it is desirable to have the control systems execute a string of such behaviors, i.e. $(\alpha_0, t_0),(\alpha_1, t_1),...,(\alpha_n, t_n)$, where $t_i~(i=1,2,...,n)$ indicates the time when the system switch from $\alpha_{i-1}$ to $\alpha_i$. Note that $\alpha_{0}$ is the original behavior with $t_0=0$. For the robotic system, $\alpha_i(x,\Gamma)|_{\Gamma=0}$ refers to the supporting polygon stay original with the dynamic characteristics similar to 4WIMD vehicles, and $\alpha_i(x,\Gamma)|_{\Gamma=1}$ means that the system adjusts the lateral distance of wheels from $d_0$ to $d_s$. Utilizing this form of behaviors, we can build upon a wealth of different control applications which all use some form of parameterized control. Therefore, the system dynamics can be simplified as
	\begin{equation}
	\label{eqmt2}
	\begin{aligned}
	\dot{x}=f_i(x,u,\Gamma) ~\textrm{for}~ t_{i} \leq t \leq t_{i+1}.
	\end{aligned}
	\end{equation}
	This model satisfies the system dynamics and the tire nonlinear effects in behavior $(\alpha_i, t_i)$. Based on the references $x^\ast$, the tracking error $e=x-x^\ast$ is defined to describe the control performance. To find the optimal control input $u$ for each behavior, the cost function is presented as
	\begin{equation}
	\label{eqmtcost}
	\begin{aligned}
	J_i=&\int_{t_{j}}^{t_{j+1}}\left(e(t)^T Q e(t)+u(t)^T R u(t)\right)dt\\
		&+x(t_{j+1})^T S x(t_{j+1}),\\
	\end{aligned}
	\end{equation}
	where $Q$, $R$, and $S$ are positive definite weight matrices. The current time is represented by $t_j$, and $t_{j+1}$ is related to the prediction horizon $\Delta{t}$ and triggering condition $\Gamma(t)$ with the following form
	\begin{equation}
	\label{eqmtj}
	t_{j+1}=\left\{
	\begin{aligned}
	&t_{i}, ~~\textrm{if}~t_{j} < t_{i} \leq t_{j}+\Delta{t}, \\
	&t_{j}+\Delta{t}, ~\textrm{otherwise}.
	\end{aligned}
	\right.
	\end{equation}
	$e(t)^T Q e(t)$ is related to the tracking error and the velocity cost of the control system. $u(t)^T R u(t)$ stands for the future value of the deviations of the model response raised by the controlled variables at the end of the prediction horizon. As the existence of $x(t_{j+1})$ for the cost function $J$, the controller drives the system state $x(t)$ at the end of the behavior $(\alpha_i, t_i)$ approaches $0$. In such condition, the stability in the switch between behaviors is ensured. The optimization problem of the controller is summarized as
	\begin{equation}
	\label{eqmt3}
	\begin{aligned}
	J &= \min_{u(t)} \sum_{i=0}^{N_p} J_i,\\
	\textrm{s.t.}~\dot{x}&=f_i(x,u,\Gamma) ~\textrm{for}~ t_{i} \leq t \leq t_{i+1},\\
	x&\in \chi,\\
	u&\in U,\\
	\end{aligned}
	\end{equation}
	where $\chi$ is limited by the wheeled robotic system, $U$ is obtained by the actuator saturation of the electric motors, and $N_p$ is the prediction horizon.  The function (\ref{eqmt3}) corresponds to the tracking ability and control volume optimization of the system. 
	
	\begin{figure}[t]
		\centering
		\color{blue}
		\includegraphics[width=1.0\hsize]{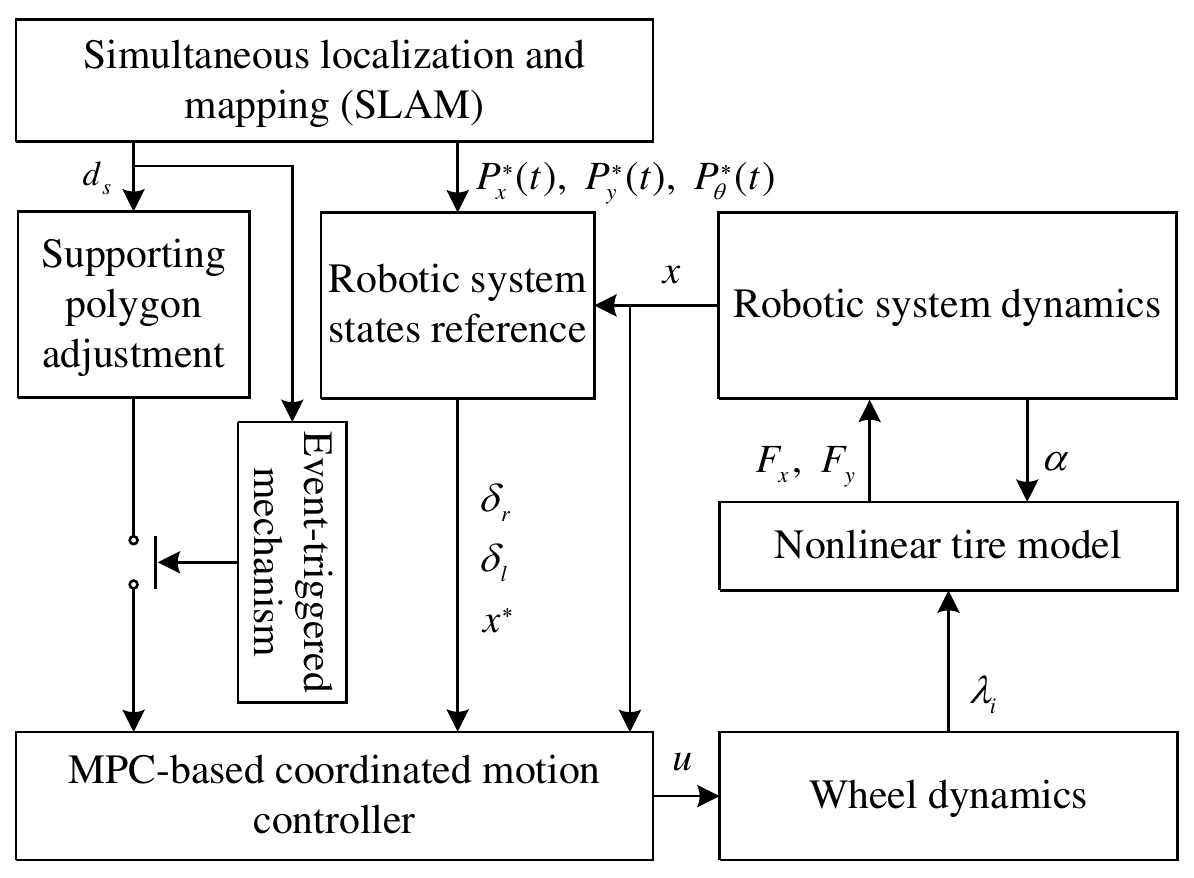}
		\caption{Block diagram of the hierarchical control system.}
		\label{fig3}
	\end{figure}
	{\color{blue} Based on the dynamics of the system, a control sequence in the next $N_p$ steps is carried out by minimizing the defined cost $J$. Then, the control sequence is used for the next $N_c$ steps, which indicates $N_p \geq N_c$. Therefore, in theory, an increasing $N_p$ leads to a rising cost (e.g., hardware resource demands or time consuming) of computing the optimization solution, and a decreasing $N_c$ improves the real-time performance and robustness.} The MPC-based coordinated motion control combining the event-based triggering mechanism is designed as shown in Fig.~\ref{fig3}. The simultaneous localization and mapping (SLAM) are implied by the information from the environment awareness system and the integrated navigation system. The obstacle information $d_s$ is sent to the event-triggered mechanism and supporting polygon adjustment to realize behaviors switch. The robotic system reference $x^\ast$ is calculated by the path planning results and the system state $x$.
	
	\section{Application and Result Analysis in Simulation and Experiment}
	
	In this section, the integrating approach is employed on a robotic system. Both the virtual prototype and the physical prototype are presented in detail.
	
	\subsection{Control System Design}
	
	\begin{figure}[!htb]
		\centering
		\includegraphics[width=1.0\hsize]{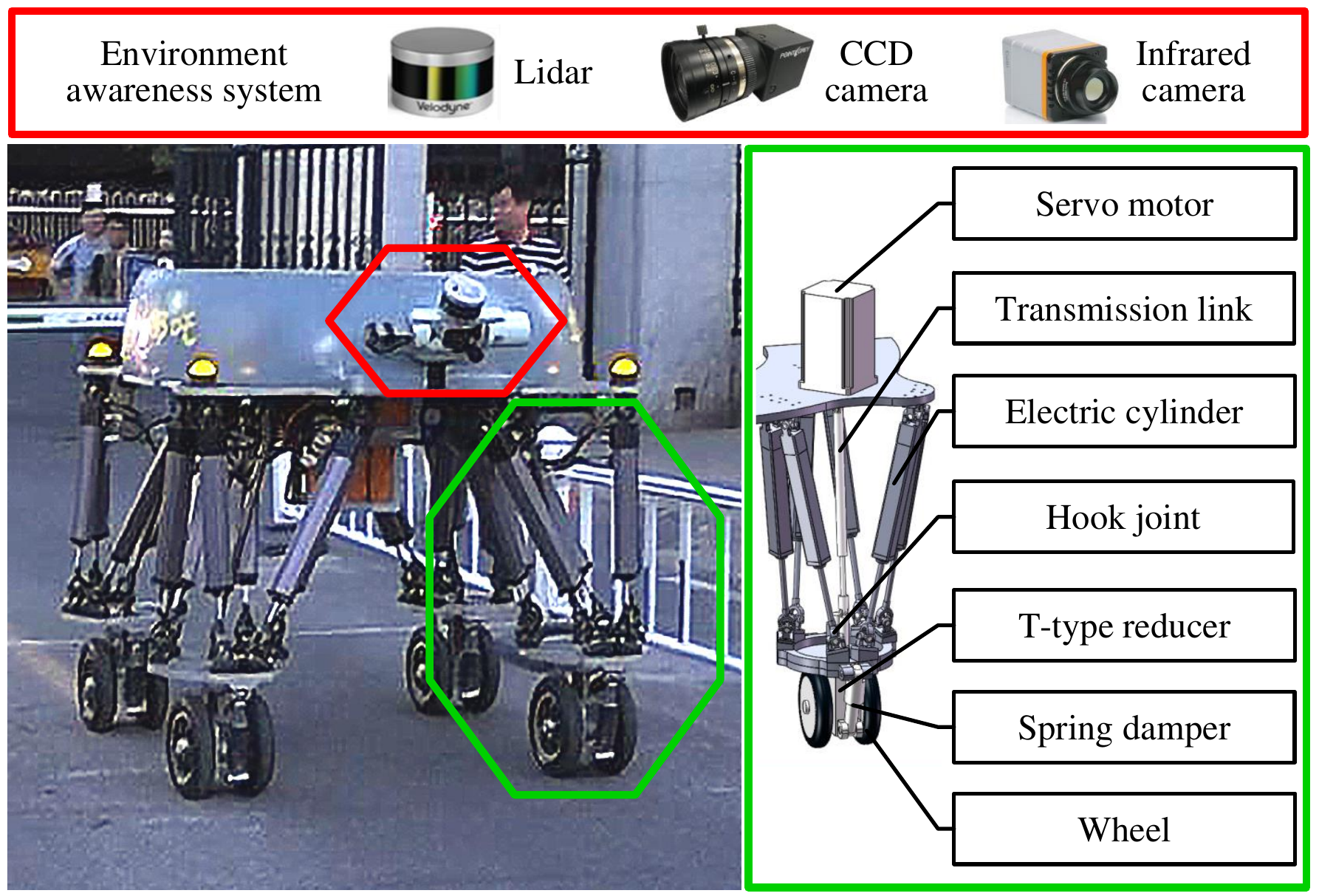}
		\caption{The physical prototype BITNAZA.}
		\label{fig4}
	\end{figure}
	\begin{table}
		\centering
		\caption{Structure Parameters of the Robotic System}
		\label{tab1}
		\begin{tabular}{cc}
			\hline
			\hline
			Parameter & Value\\
			\hline
			Robot Length*Width*Height (m) & 1.5*1.5*1.2 \\
			Robot Weight (kg) & 278.0 \\
			Tire Diameter (m) & 0.2\\
			Battery Life (h) & 2.5 \\
			Carrying Capacity (kg) & 300.0\\
			Reducer Reduction Ratio & 40:1 \\
			Maximum Angle of Climb ($^\circ$) & 27.0\\
			Maximum Velocity (km/h) & 10.0\\
			\hline
			\hline
		\end{tabular}%
		\label{tab:addlabel}%
	\end{table}%
	
	We have established a four wheel-leg independent motor-driven robotic system with a dynamic supporting polygon. The physical prototype named BITNAZA is shown in Fig.~\ref{fig4}, and the structure parameters of the robotic system in listed in Table~\ref{tab:addlabel}. The robotic system consists of four Stewart structured wheel-legs, the control system, the environment awareness system, the integrated navigation system, and the power system. The wheel-leg is presented in detail inside the green box. A transmission link and a T-type reducer make up the transmission system of the wheel-leg. The hook joints and the spring dampers connect the other components. Driven by 6 electric cylinders and a servo motor, the wheels can realize translations and rotations in 6 degrees of freedom. As a result, the BITNAZA can not only move and steer like a 4WIMD system but also adjust the supporting polygon meeting the dynamic model in Fig.~\ref{fig1}. The environment awareness system is introduced in the red box, including a lidar, a charge-coupled device (CCD) camera, and an infrared camera. Equipped with a compact, portable, and tightly coupled synchronized position attitude navigation (SPAN-CPT), the integrated navigation system bring together two different, but complementary technologies: global navigation satellite system (GNSS) positioning and inertial navigation. Therefore, the robotic system achieves all-weather path planning and obstacle detection.

\begin{figure}[t]
		\centering
		\includegraphics[width=1.0\hsize]{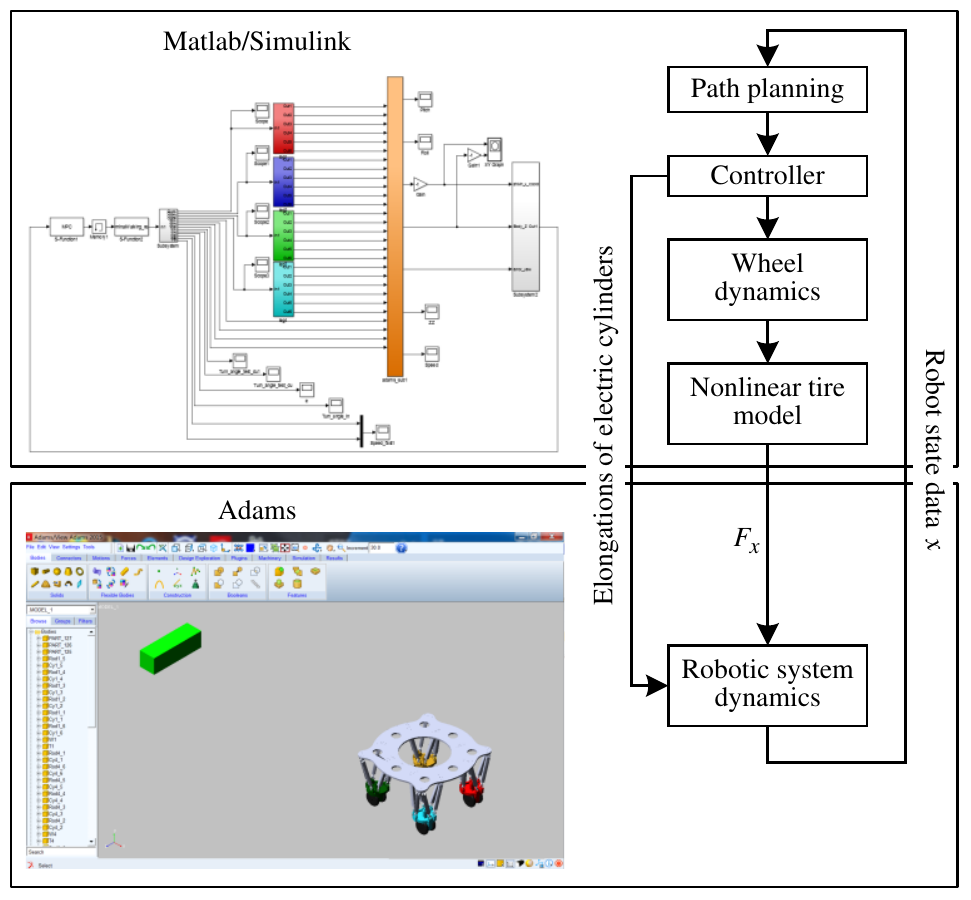}
		\caption{The co-simulation platform.}
		\label{fig5}
	\end{figure}

	To improve the safety and efficiency of designing and testing, a co-simulation platform is built as shown in Fig.~\ref{fig5}. The proposed controller is implied by Matlab/Simulink and drives the virtual prototype in Adams, while the state data of the prototype are fed back to the controller in real-time. The control inputs of the virtual prototype are $F_x$ and the elongations of electric cylinders, while the wheel dynamics and nonlinear tire model are concerned in the Matlab/Simulink. Note that the virtual prototype ignores some factors such as the friction in the electric cylinders and the hook joints.
	
	\subsection{Simulation Results}
	
	To evaluate the coordinated control performance of this robotic system, a series of comparative simulation tests are carried out. Test 1 and Test 2 stands for the tracking performance of the robotic system controlled by the proposed algorithm with different parameters. For controller No.1, the control horizon $N_c=30$ and prediction horizon $N_p=60$ are implied. The decreased $N_c=5$ and $N_p=20$ are set in controller No.2. The reference Line 1 is a single lane change that is widely used to test the tracking ability for wheeled systems. The road is flat without any obstacle so that the supporting polygon of the robotic system stays static. {\color{blue} In Test 3 and Test 4, the controllers are the same. However, there exist two obstacles on the road in Test 4, while no obstacle is placed on the road in Test 3.} For clarity and intuition, Table~\ref{tab:test} is organized.
	
	\begin{figure}[t]
		\centering
		\includegraphics[width=1.0\hsize]{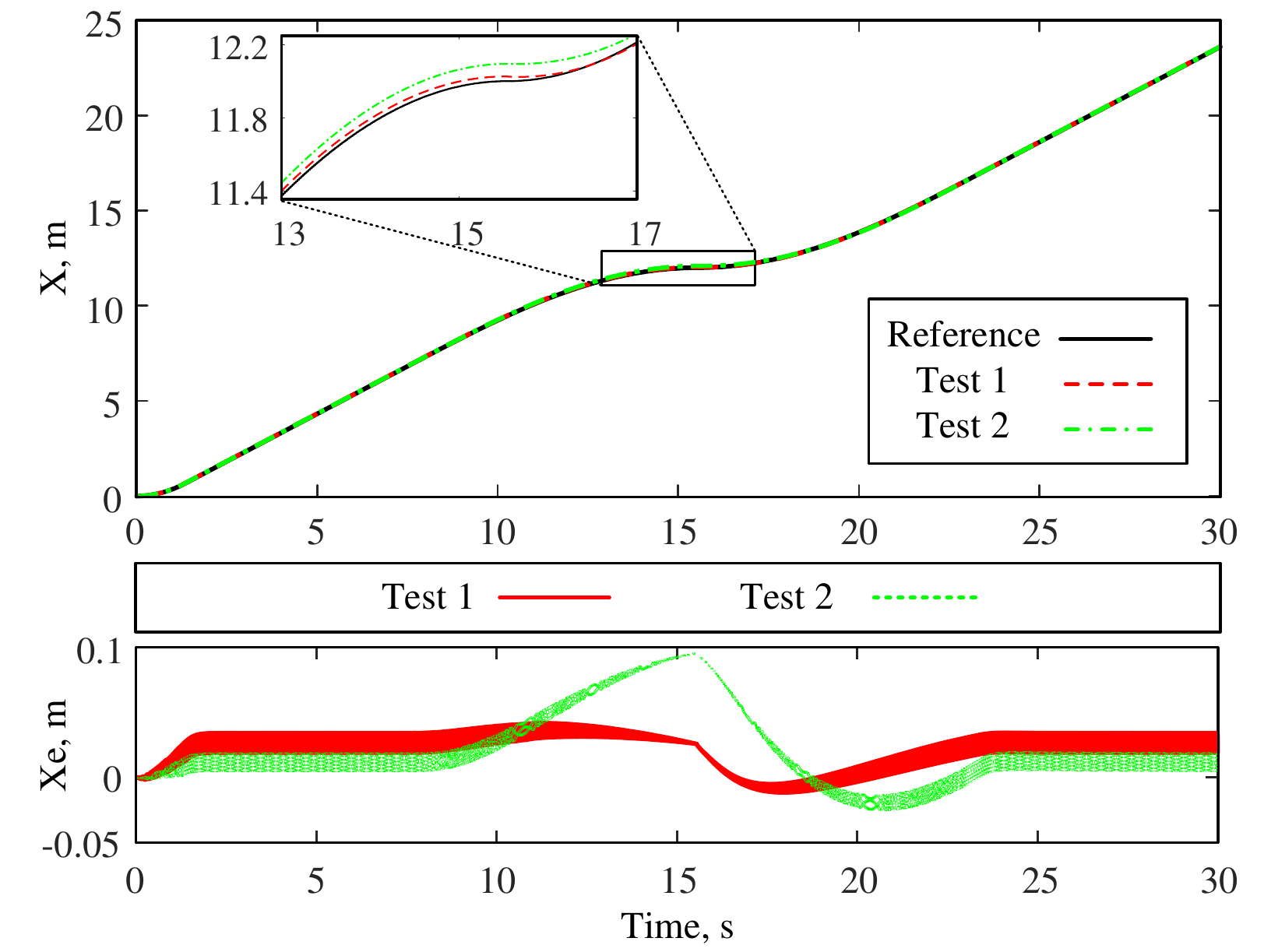}
		\caption{Tracking curves and error curves in X-axis for Test 1 and Test 2.}
		\label{fig6}
	\end{figure}
	\begin{figure}[!htb]
		\centering
		\includegraphics[width=1.0\hsize]{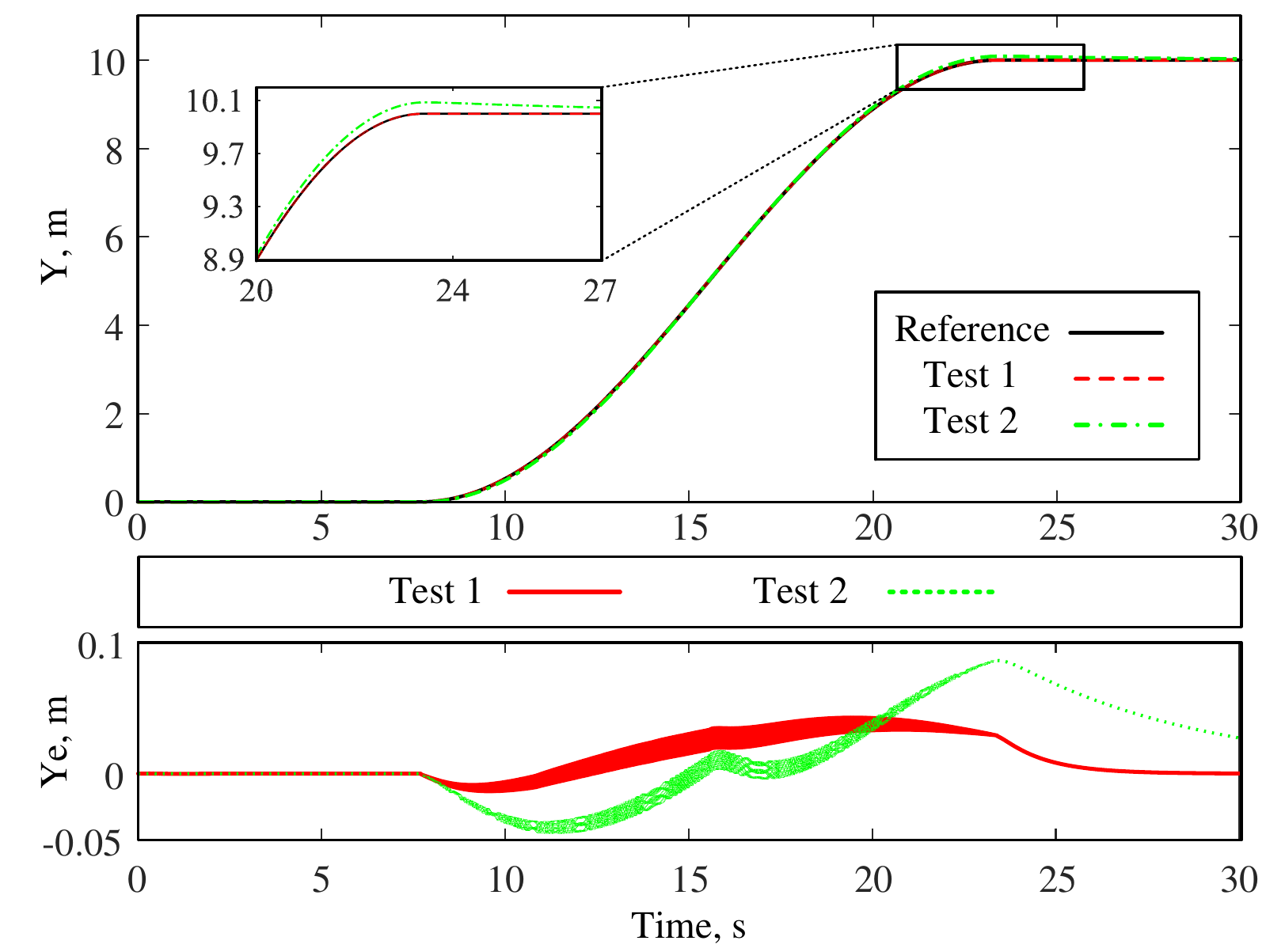}
		\caption{Tracking curves and error curves in Y-axis for Test 1 and Test 2.}
		\label{fig7}
	\end{figure}
	\begin{figure}[!htb]
		\centering
		\includegraphics[width=1.0\hsize]{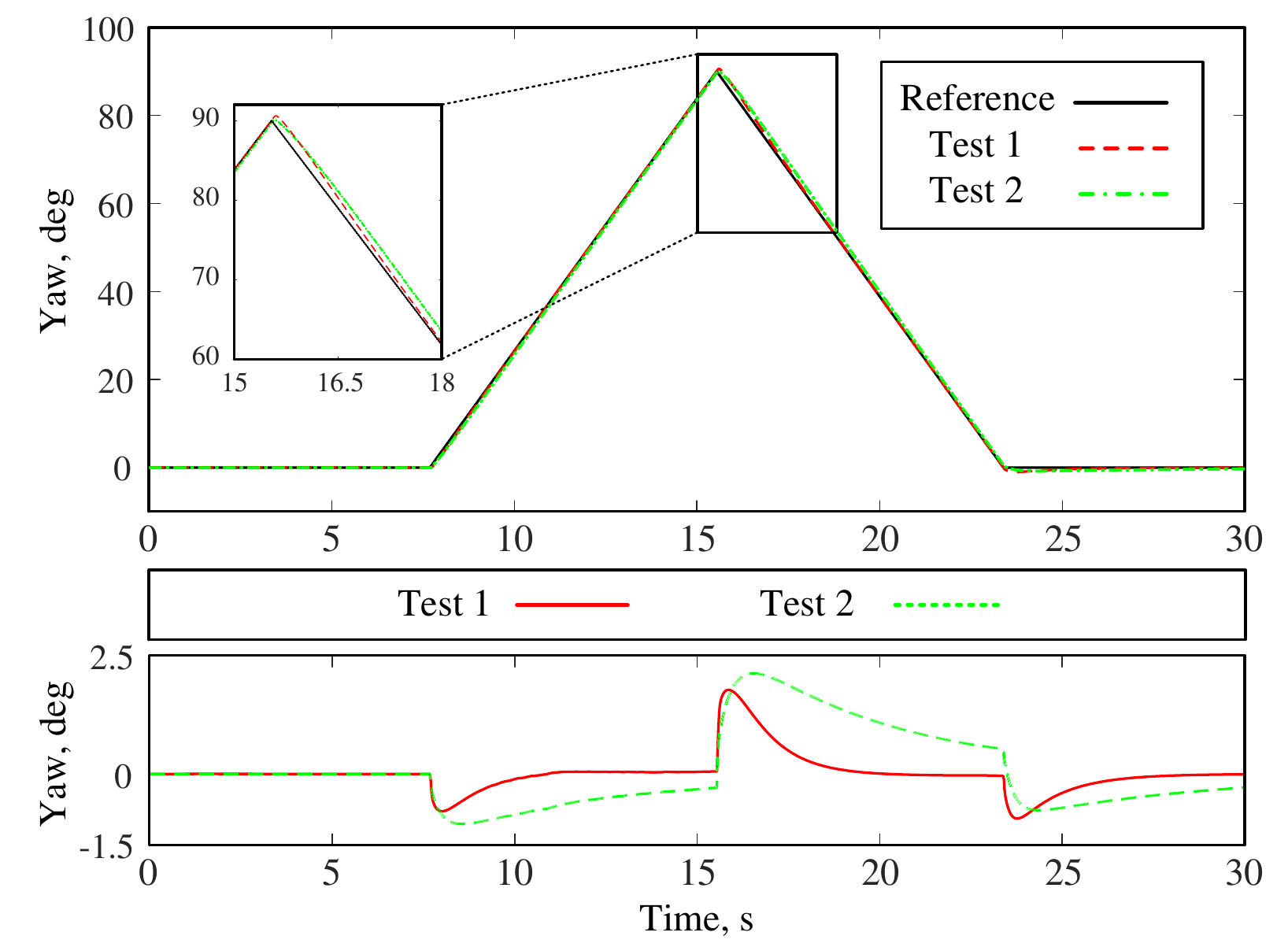}
		\caption{Tracking curves and error curves of Yaw for Test 1 and Test 2.}
		\label{fig8}
	\end{figure}
	\begin{figure}[!htb]
		\centering
		\includegraphics[width=1.0\hsize]{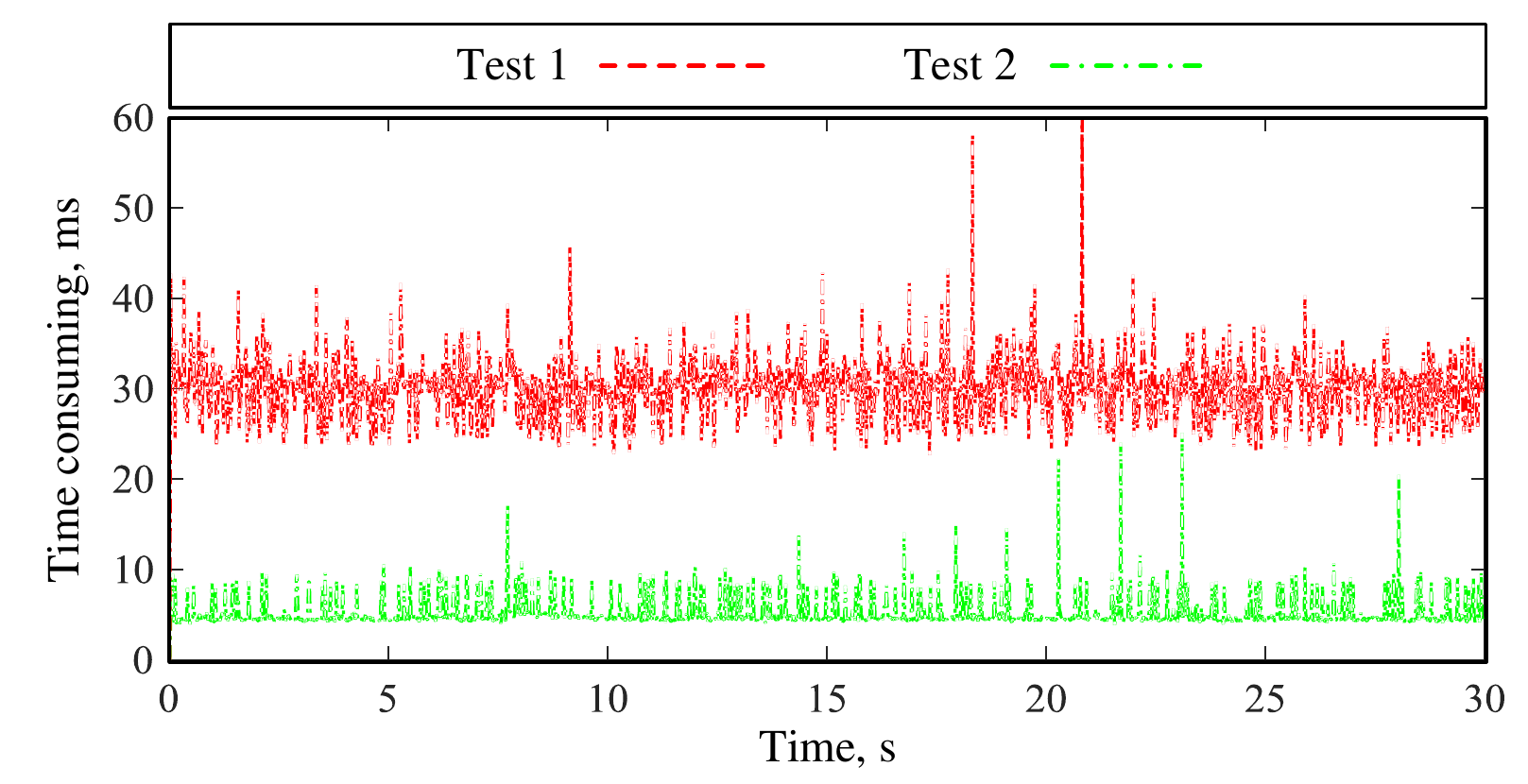}
		\caption{The time consuming of the controllers for Test 1 and Test 2.}
		\label{fig9}
	\end{figure}
	
	\begin{table}[!htb]
		\centering
		\caption{Evaluation Procedures}
		\begin{tabular}{cccc}
			\hline
			\hline
			Test & Trajectory & Obstacle & Controller Number\\
			\hline
			Test 1 & Line 1 & N & No.1 \\
			Test 2 & Line 1 & N & No.2 \\
			Test 3 & Line 2 & N & No.1 \\
			Test 4 & Line 2 & Y & No.1 \\
			\hline
			\hline
		\end{tabular}%
		\label{tab:test}%
	\end{table}%
	\begin{figure*}[t]
		\centering
		\includegraphics[width=1.0\hsize]{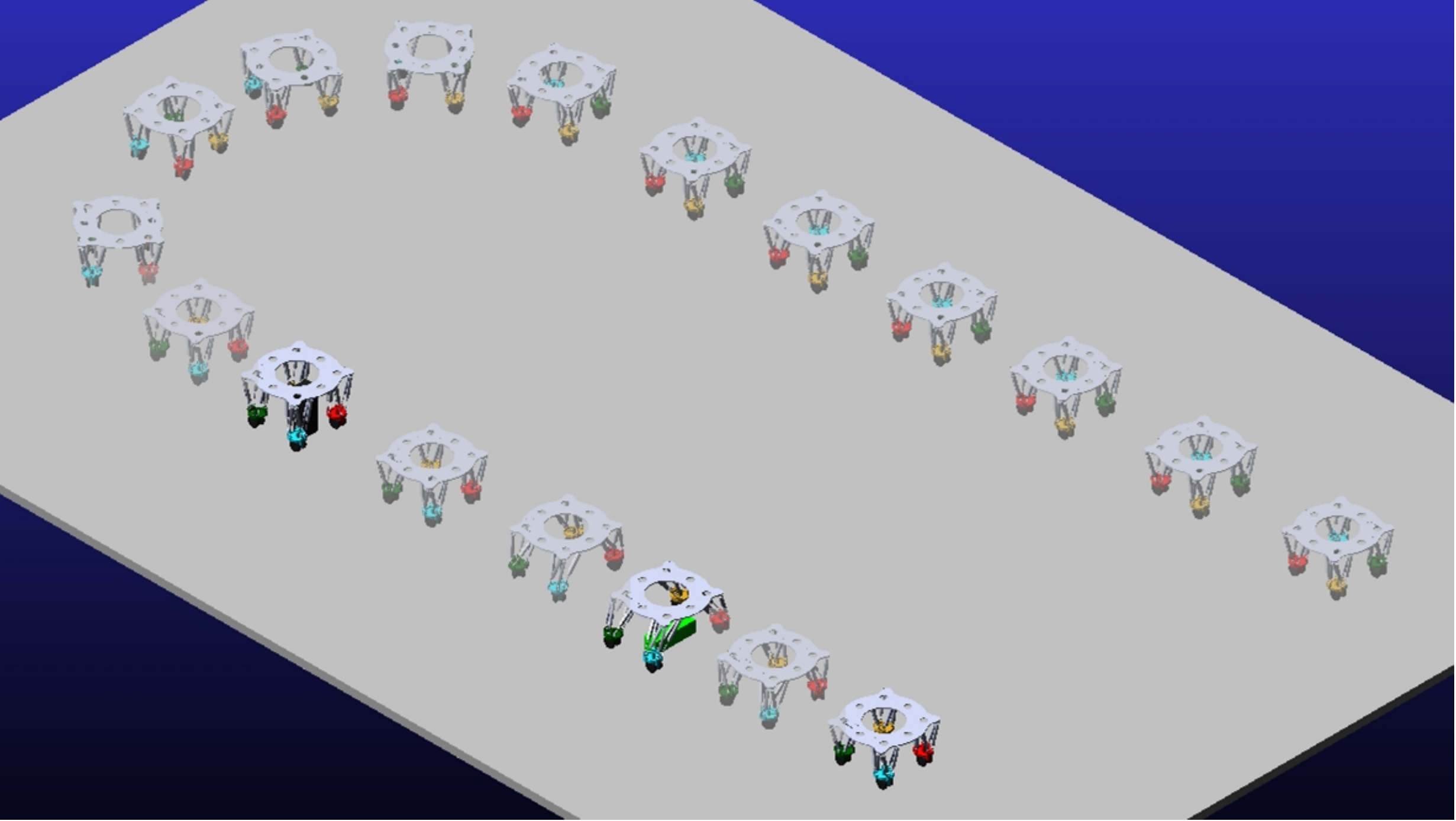}
		\caption{The result recording of Test 4.}
		\label{figsimulation}
	\end{figure*}
	\begin{figure}[!htb]
		\centering
		\includegraphics[width=1.0\hsize]{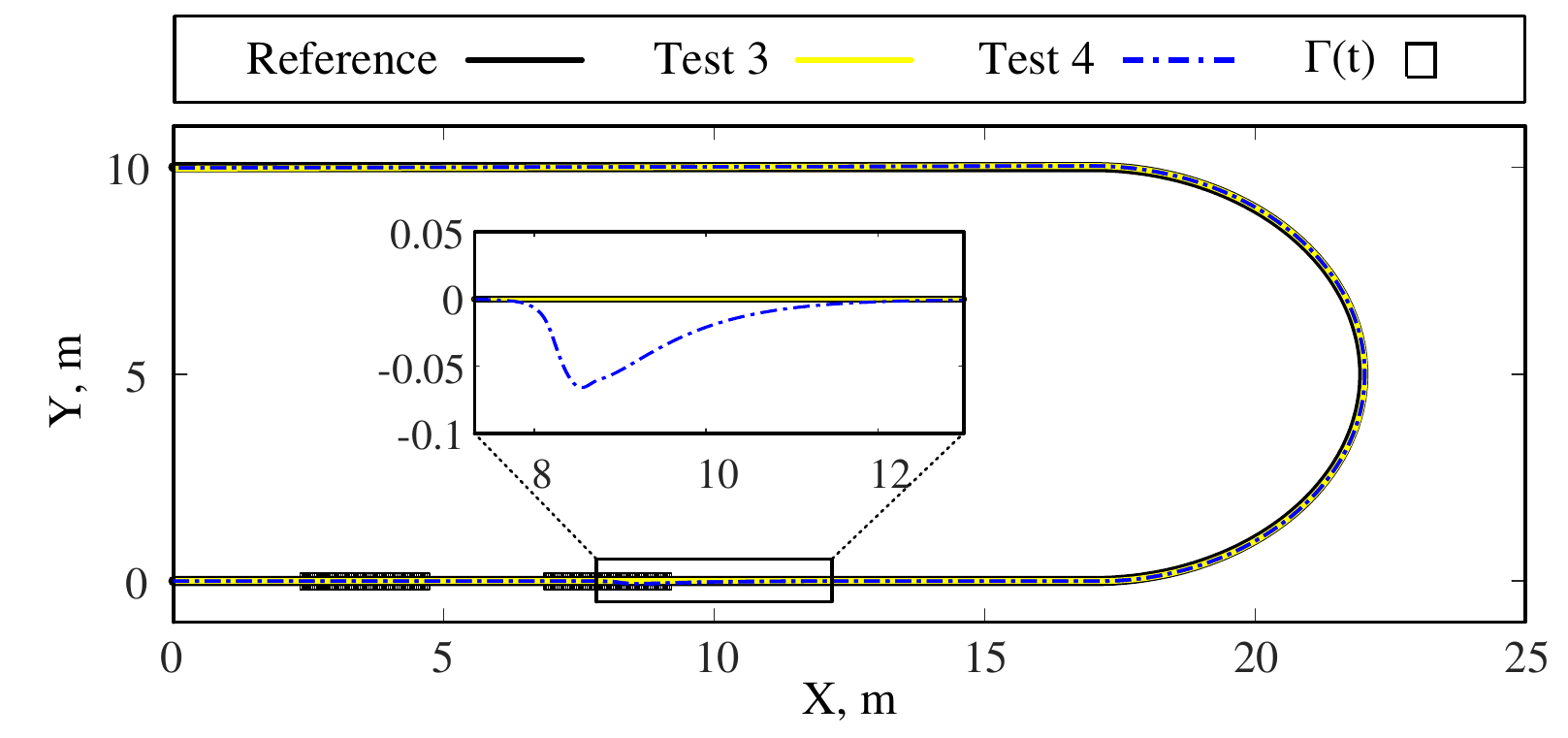}
		\caption{Tracking curves for Test 3 and Test 4.}
		\label{fig10}
	\end{figure}

	Considering the parameters in Table~\ref{tab:addlabel} and the system model, the weight matrix are chosen as $Q=\textrm{diag}(1.0, 10.0)$ and $R=\textrm{diag}(1.0, 1.0, 1.0, 1.0)$. The constraint of the input $u$ is
	\begin{equation}
	\label{eqmtest1}
	U=\left[
	\begin{array}{cc}
	-10.0 & 10.0 \\
	-10.0 & 10.0 \\
	-10.0 & 10.0 \\
	-10.0 & 10.0 \\
	\end{array}
	\right].
	\end{equation}
	
	The simulation results of Test 1 and Test 2 are shown as follows. In Fig.~\ref{fig6}, X indicates the position of the robot centroid in the X-axis and Xe implies the tracking error in X-axis, which are similarly defined in Fig.~\ref{fig7} and Fig.~\ref{fig8}. The maximum tracking errors in X-axis, Y-axis, and Yaw are 0.042 m, 0.043 m, and $1.78^\circ$ respectively for Test 1, while the same indicators are 0.095 m, 0.086 m, and $2.13^\circ$ for Test 2. Therefore, with the increments of $N_p$ and $N_c$, tracking accuracy is improved for the proposed algorithm.
	
	Moreover, as presented in Fig.~\ref{fig9}, the time consuming for each cycle of the controller is measured. Controller No.1 needs 30 ms for each cycle of calculation on average, while controller No.2 only needs 7.5 ms approximately. Therefore, with the increments of $N_p$ and $N_c$, the time consumption is also increased, which may limit the application of the proposed algorithm for real-time motion control systems. This feature indicates that the $N_p$ and $N_c$ must be reasonably chosen for our robotic system in practice.
	
	Since the hardware of the simulation system is sufficient, controller No.1 is adopted in the following simulation tests. Test 4 is recorded in Fig.~\ref{figsimulation}, where there are two different obstacles. The first obstacle in green is 1.6 m in width and 0.4 m in height. The second obstacle in black is 0.5 m in width and 1.45 m in height. Considering the structure parameters in Table~\ref{tab:addlabel}, the robotic system can adjust the lateral wheel-track or the body height to overcome the obstacles. From the analysis in Fig.~\ref{figsimulation}, the system dynamic model has the same characteristics with different body heights. As shown in (\ref{eqpf1}), it holds $\Gamma(t)=1$ in the process of the lateral wheel-track adjustment.
	
	The tracking curves of the robotic system in Test 3 and Test 4 are as shown in Fig.~\ref{fig10}. Note that the small black box is set when $\Gamma(t)=1$. It is pointed out there exists a peak of up to -0.074 m of the tracking error in the second process of the lateral wheel-track adjustment. {\color{blue}However, the tracking error of the system with controller No.1 converges quickly. On the other hand, the small black box occurs only in the process of supporting polygon adjustment. The event-triggering mechanism efficiently splits the control model into piecewise functions by behaviors. The system function is linear when the robot tracking the desired path with $\Gamma(t)=0$. In such a situation, the proposed controller helps to save hardware resources and energy. }
	
	The simulation results presented above prove that the proposed control algorithm is stable and effective. {\color{blue}On the other hand, the different performances in Test 1 to Test 4 guide the parameter tuning of the controller in practical applications.}
	
	\subsection{Experiment Results}
	
\begin{figure}[t]
		\centering
		\includegraphics[width=1.0\hsize]{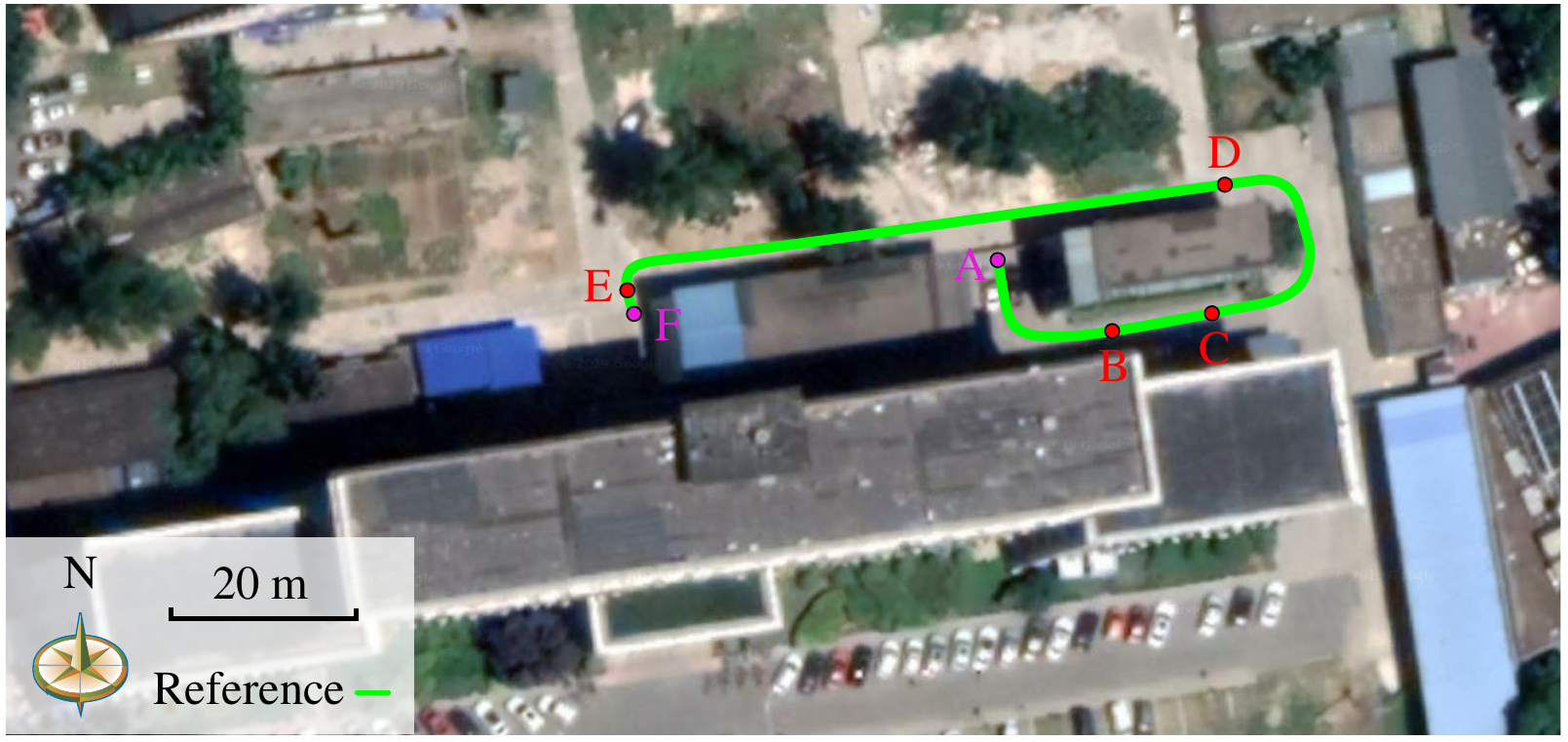}
		\caption{Experimental environment and the reference trajectory on map.}
		\label{fig12}
	\end{figure}

\begin{figure*}[t]
		\centering
		\includegraphics[width=1.0\hsize]{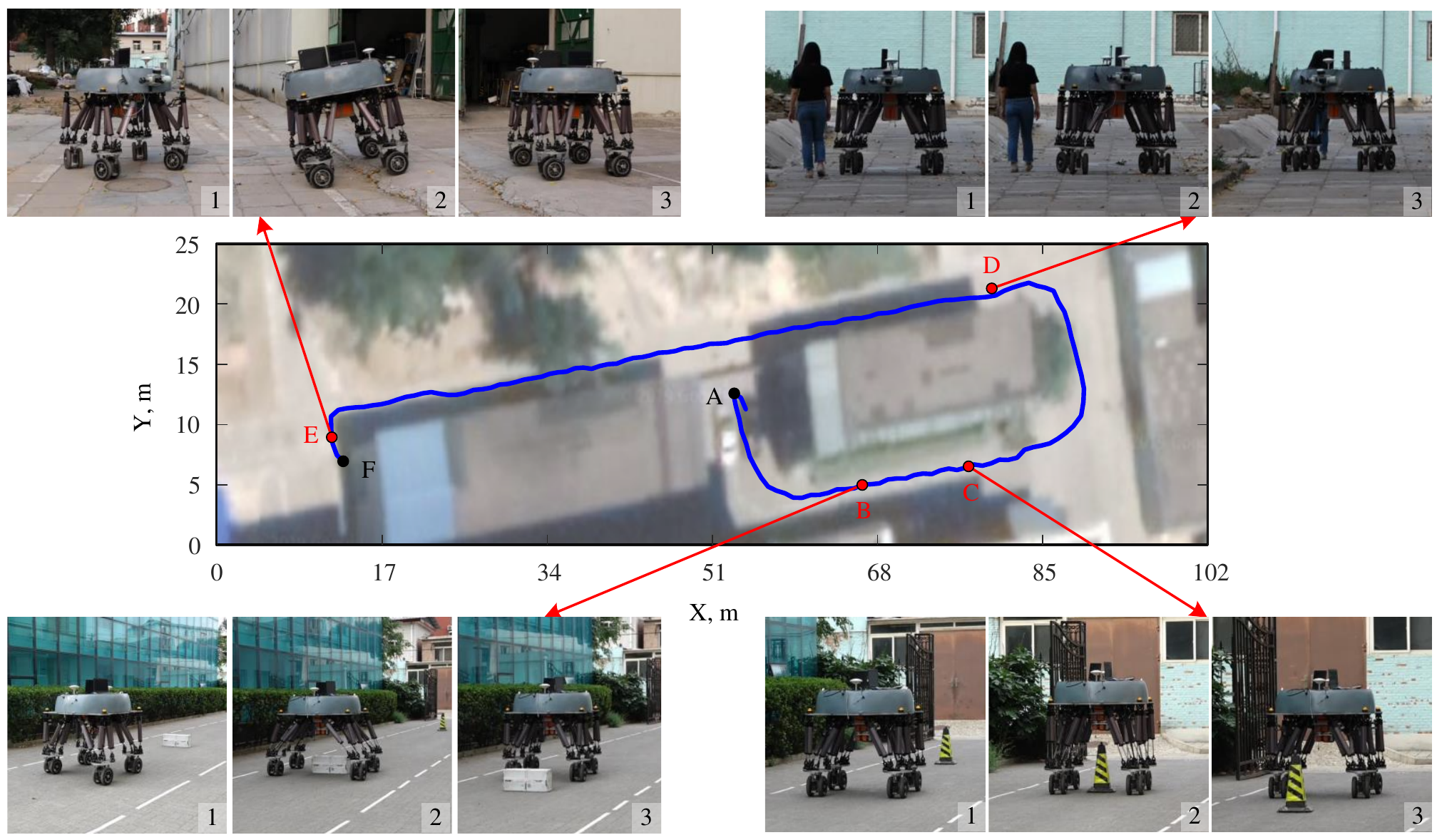}
		\caption{Tracking curve for the experiment.}
		\label{fig13}
	\end{figure*}

	To estimate the coordinated motion control performance, the naturalistic driving data are collected in $39.970^\circ$ E and $116.316^\circ$ N. Some cases that include straight, turning, passing intersections, bypassing pedestrians, and crossing obstacles are set at markers $A$ to $F$ to analyze the efficiency of the event-based MPC controller. The route is shown in Fig.~\ref{fig12}, covering a distance of approximately 134 m. In the experiment, the SLAM is carried out by the environment awareness system and the integrated navigation system as shown in Fig.~\ref{fig4}. A large number of experiments are conducted with various obstacles. One case from the naturalistic driving database is presented and analyzed in Fig.~\ref{fig13}. The processes at the marked points are recorded in detail.


	The experiment starts at marker $A$ followed by a short straight road and a left turn. At marker $B$, a white obstacle, which is 1.5 m in width, is set. A black and yellow obstacle appears at marker $C$, which is 1.3 m in height. The triggering condition $\Gamma(t)= 1$ holds while the robot is crossing the obstacle at marker $B$ and adjusts the supporting polygon, while $\Gamma(t)= 0$ holds at marker $C$ with the robot overcoming the obstacle. A pedestrian and a slope are placed at markers $D$ and $E$. As the pedestrian is 1.65 m in height which is more than the maximum height, the robotic system avoids the pedestrian by passing from {\color{blue}the side}. The angle of the slope is 16.5$^\circ$ which is smaller than the maximum angle of climb, the robotic system climbs up the slope. The robot smoothly passes the different terrains tracking the desired path from the SLAM. The experiment ends at the marker $F$ and the results of the physical prototype trajectory are shown in Fig.~\ref{fig13}. {\color{blue}The experiment takes about 2.5 minutes entirely, such that the average cruise speed approximates 0.9 m/s.}
	
	During the control process, the motor torques are distributed according to the driving conditions of the control system. There is no collision with obstacles or slippage. From the experiment results, it can be concluded that the proposed controller has significant performance in coordinated motion control for the {\color{blue}four-wheel-leg} independent motor-driven robotic systems.

	\section{Discussion}
	
	This paper studies a coordinated motion control and obstacle-crossing problem for the four wheel-leg independent motor-driven robotic systems via the model predictive control (MPC) approach based on an event-triggering mechanism. The wheel-leg robotic system with a dynamic supporting polygon is analyzed and modeled. The system model are proposed in detail, and the single wheel dynamics considering the characteristics of motor-driven and the Burckhardt nonlinear tire model. {\color{blue}The relative position changes between wheels result in a nonlinear time-varying model for the control system, which brings the main challenge for MPC controller design. As a matter of the fact that the supporting polygon is only adjusted at certain conditions, an event-based triggering mechanism is designed to split the control model into piecewise functions. When the robot requires no more supporting polygon adjustment, the system function is time-invariant, such that the optimal solution of the MPC controller is simplified to save hardware resources and energy. Based on the developed cost function, the controller drives the system state at the end of the behavior approaches $0$, which guarantees the switching stability between behaviors.} The simulated and experimental results verify the efficiency of the proposed controller. 
	
	\biboptions{numbers,sort&compress}	
	\bibliographystyle{IEEEtran}
	\bibliography{mybibfile}
	
\end{document}